\documentclass{article}

\usepackage[preprint]{neurips_2025}  

\usepackage[utf8]{inputenc} 
\usepackage[T1]{fontenc}    
\usepackage[hidelinks]{hyperref}       
\usepackage{url}            
\usepackage{booktabs}       
\usepackage{amsfonts}       
\usepackage{nicefrac}       
\usepackage{microtype}      
\usepackage{xcolor}         
\usepackage{hyperref}

\usepackage{amsmath}    
\usepackage{amssymb}    
\usepackage{bm}         
\usepackage{mathtools}  
\usepackage{nccmath}
\usepackage{arydshln}
\usepackage{multirow}

\usepackage{graphicx}
\usepackage{subcaption} 

\DeclarePairedDelimiter{\nint}\lfloor\rceil

\title{QwT-v2: Practical, Effective and Efficient Post-Training Quantization}

\author{%
  Ningyuan Tang \quad \quad Minghao Fu \quad \quad Hao Yu \quad \quad Jianxin Wu\thanks{Corresponding author.} \\
  National Key Laboratory for Novel Software Technology, Nanjing University, China\\
  School of Artificial Intelligence, Nanjing University, China\\
  \texttt{tangny@lamda.nju.edu.cn, fumh@lamda.nju.edu.cn, wujx2001@nju.edu.cn} \\
}

\begin{document}

\maketitle

\begin{abstract}
Network quantization is arguably one of the most practical network compression approaches for reducing the enormous resource consumption of modern deep neural networks. They usually require diverse and subtle design choices for specific architecture and tasks. Instead, the QwT method is a simple and general approach which introduces lightweight additional structures to improve quantization. But QwT incurs extra parameters and latency. More importantly, QwT is not compatible with many hardware platforms. In this paper, we propose QwT-v2, which not only enjoys all advantages of but also resolves major defects of QwT. By adopting a very lightweight channel-wise affine compensation (CWAC) module, QwT-v2 introduces significantly less extra parameters and computations compared to QwT, and at the same time matches or even outperforms QwT in accuracy. The compensation module of QwT-v2 can be integrated into quantization inference engines with little effort, which not only effectively removes the extra costs but also makes it compatible with most existing hardware platforms.
\end{abstract}

\section{Introduction}
The rapid scaling of deep neural networks guarantees their strong performance in various fields, including language (\cite{bert,gpt3}), vision (\cite{vit,mask_rcnn}), speech (\cite{speechgpt}), and multimodal (\cite{CLIP,llava}), but also incurs heavy burden on hardware resources such as compute, storage and energy. Network quantization (\cite{quantization_survey}) stands out as a highly practical approach to reduce these resource consumptions, highlighting large compression ratio, low accuracy loss, and high hardware friendliness. Existing quantization methods are often categorized into Quantization-Aware Training (QAT) which requires training, and Post-Training Quantization (PTQ) which is training free.

Neither PTQ nor QAT is perfect. PTQ (\cite{brecq,smooth}) methods are fast but may lead to considerable accuracy degradation when the bit-width is low, while QAT (\cite{LSQ,zhu2023quantized,Q-ViT}) methods are thousands of times slower than PTQ methods. Moreover, existing quantization methods require meticulous design for different models and tasks. Changes in model architecture or task requirements can potentially render these quantization methods ineffective. For example, the issue of handling outliers is not important in quantizing CNNs, but is critical for quantizing Transformers.

Quantization without Tears (QwT) by~\cite{QwT} is a simple and general quantization method which suits various network architectures and applications. By adding external compensation matrices to a quantized network $S^\mathbb{Z}$ (i.e., the output of a PTQ method), QwT can effectively improve network accuracy without severely affecting latency or model size. In addition, parameters of compensation modules in QwT can be set in closed-form with a small calibration set. This fact enables QwT to train very quickly without any hyperparameter tuning.

But, despite being simple, effective and widely applicable, a few drawbacks in QwT remain, which critically limit its deployment in many other real-world scenarios.
\begin{enumerate}
	\item \textbf{Hardware compatibility.} QwT requires the inference engine to compute mainly in quantized fixed-point format (e.g., INT8), but there are also floating-point computations (e.g., FP16 or BF16) for the compensation modules. Although this capability is available in many GPUs, it is not yet supported well in many other hardware platforms (e.g., simple AI accelerators in a cellphone or an embedded hardware) or software inference engines like vLLM (\cite{kwon2023efficient}). This incompatibility severely restricts QwT's usability.
	\item \textbf{Extra parameters.} QwT compensates a PTQ model with linear projections. In high-bit quantization, compensation matrices only account for a small percentage of the network size. But in a low-bit quantized network, the floating-point compensation matrices become non-negligible and increase model size by around 30\%, as shown by~\citet{QwT}.
    \item \textbf{Lower efficiency.} For each compensated block, QwT receives the block input, calculates floating-point compensation, and adds it to the block output. During this process, the inference engine needs to store all these tensors, which increases both memory footage and latency (due to both more memory accesses and more computations).
\end{enumerate}

In this paper, we show that all these issues can be resolved simultaneously. We propose QwT-v2, which inherits the advantages of QwT and also showcases the following improvements over it:
\begin{itemize}
	\item \textbf{Minimal extra parameter \& computation.} We introduce a new form of compensation module, named channel-wise affine compensation (CWAC), which requires significantly less parameters and computations than those in QwT.
	\item \textbf{Easy integration \& no floating-point computation.} Furthermore, the CWAC module can be integrated into inference engines and then it incurs zero overhead during inference. The integration requires only few efforts---merely changing quantization parameters without changing the computation graph. After this integration, floating-point computations are \emph{not} required, hence QwT-v2 \emph{suits most hardware platforms}.
	\item \textbf{Fine-grained compensation.} Thanks to the parameter and computation efficiency of CWAC, we can compensate each quantized module individually, rather than compensating for an entire network block. This guarantees better result of QwT-v2 in many settings.
\end{itemize}

\section{Related Works}

As aforementioned, network quantization methods can be categorized into two lines: Quantization-Aware Training (QAT) and Post-Training Quantization (PTQ). 

QAT methods (\cite{ReactNet,ewgs,LSQ,LSQ+,Q-ViT}) simulate low-bit computation during training, and minimize accuracy degradation caused by quantization. LSQ (\cite{LSQ}) updates not only parameters but also scale factors via gradient descent during training. LSQ+ (\cite{LSQ+}) extends LSQ to non-uniform quantization. Q-ViT (\cite{Q-ViT}) improves quantization performance on Transformers by rectifying information distortion of quantized self-attention maps.

PTQ methods (\cite{brecq,PTQ4ViT,gptq}) convert a well-trained floating-point model to low-bit fixed-point one with a small calibration set. Several methods (\cite{adaround,brecq,qdrop}) work well on CNNs but fail to be effective on Transformers, due to different network structure and weight/activation numerical distributions. Methods have also been proposed to quantize Transformers (\cite{PTQ4ViT,gptq,RepQ-ViT,FQ-ViT,smooth}). FQ-ViT (\cite{FQ-ViT}) uses power-of-2 factor and log-int-softmax quantization to tackle the difficulty of quantizing layer norm and softmax activations. RepQ-ViT (\cite{RepQ-ViT}) transforms complex quantizers to simple quantizers via scale reparameterization for efficient inference. GPTQ (\cite{gptq}) exploits a single-shot weight quantization technique which utilizes approximate second-order information.

\section{The QwT-v2 Method}

QwT by \cite{QwT} proposes a simple and general quantization method to improve PTQ methods, and it is effective on both CNNs and Transformers. For example, given a Transformer block (i.e., MHSA+FFN) whose output in the floating-point and quantized networks are $Y^{\mathrm{full}}$ and $Y^{\mathrm{quant}}$, respectively; while its input in the quantized version is $X^\mathrm{quant}$. QwT noticed that quantization leads to the difference between $Y^{\mathrm{full}}$ and $Y^{\mathrm{quant}}$, and reducing this difference helps the quantized model a lot. It linearly regresses $Y^{\mathrm{full}}-Y^{\mathrm{quant}}$ using $X^\mathrm{quant}$, whose regression parameters $(W,b)$ can be obtained in closed-form. Then, a compensation term $W X^\mathrm{quant} + b$ is implemented as a fully connected (FC) layer and is added to $X^\mathrm{quant}$, such that $Y^{\mathrm{quant}}+W X^\mathrm{quant} + b$ (compared to $Y^{\mathrm{quant}}$ alone) is closer to $Y^{\mathrm{full}}$. The main idea of QwT is illustrated in the top half of Figure~\ref{fig:mot_a}.

\subsection{Key Ideas in Our QwT-v2}

QwT improves PTQ methods' accuracy significantly. But, since $W X^\mathrm{quant} + b$ is computed in floating-point format while other computations are in fixed-point (quantized) format, QwT is not compatible with some hardware platforms. Furthermore, the existence of $W$ adds roughly 30\% model size, together with computations and memory accesses to the quantized model. Our goal is to design a PTQ method that suits almost all hardware, is parameter efficient, and also memory efficient.

\begin{figure}
    \centering
    \begin{subfigure}[b]{0.33\linewidth}
        \includegraphics[width=\linewidth]{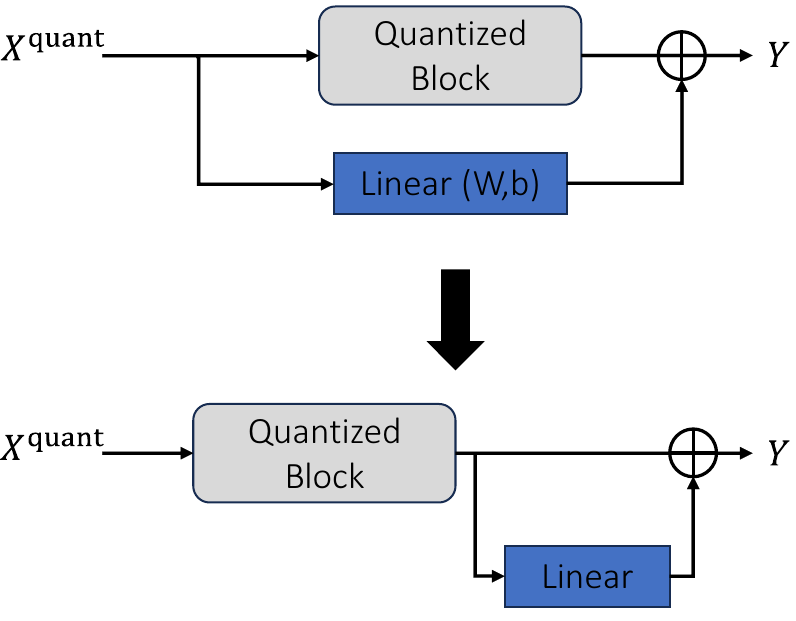}
        \caption{Pre- \& post-compensation} 
        \label{fig:mot_a}
    \end{subfigure}
    \hfill
    \begin{subfigure}[b]{0.66\linewidth}
        \includegraphics[width=\linewidth]{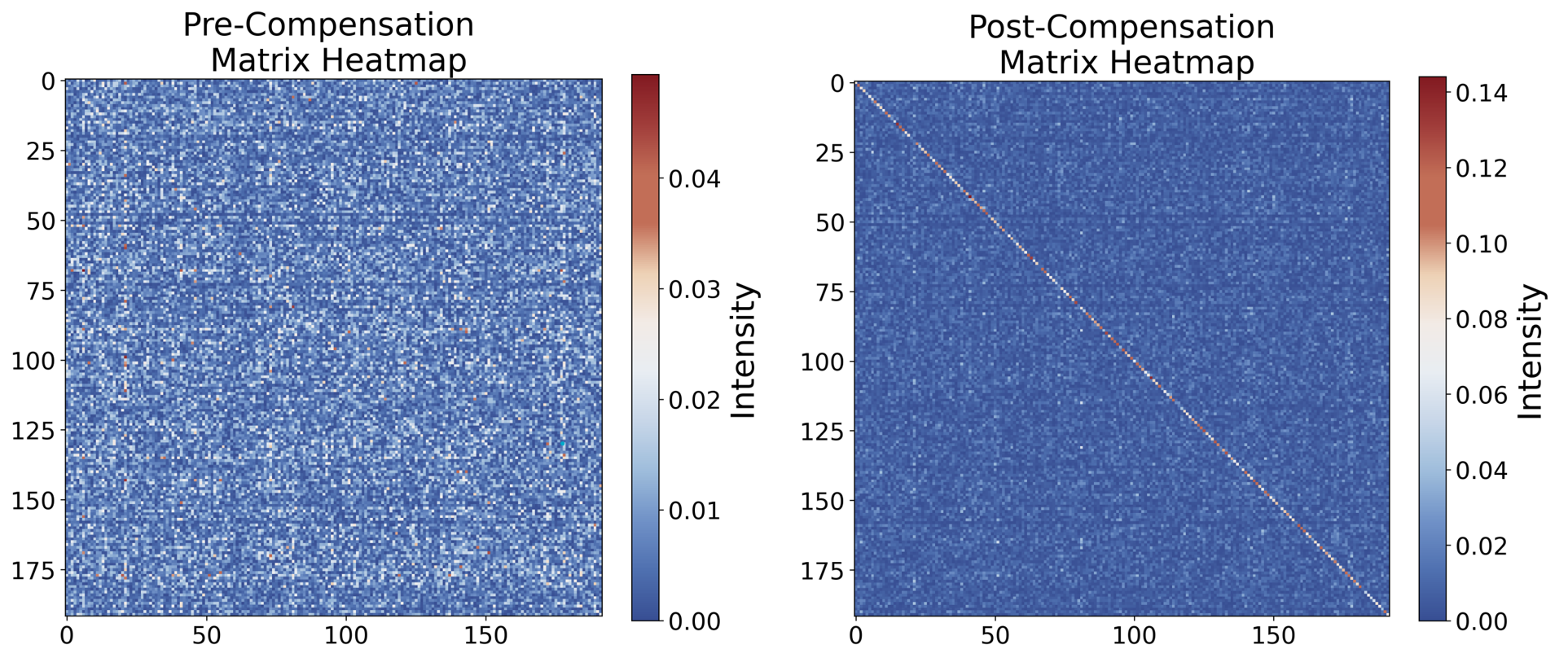}
        \caption{Example distribution of values in $W$} 
        \label{fig:mot_b}
    \end{subfigure}
    \caption{The left figure~(\subref{fig:mot_a}) shows pre-compensation in QwT (top) and our post-compensation in QwT-v2 (bottom), respectively. In pre-compensation, the linear compensation module receives a Transformer block's input as its input, while in post-compensation, the block's output is input to the linear compensation module. The right figure~(\subref{fig:mot_b}) shows example compensation matrix $W$'s absolute values in QwT and our QwT-v2, respectively (on 4-bit quantized DeiT-tiny).}
    \label{fig:mot}
\end{figure}

We start by asking whether QwT is the most suitable compensation method. QwT \emph{linearly} compensates \emph{an entire Transformer block}'s output with its input, but such a block involves high \emph{non-linearity} which cannot be easily modeled by a linear projection. We should \emph{try to minimize the non-linearity} by changing the regressor's input form the Transformer block's input to its output, as shown in Figure~\ref{fig:mot_a}. We call them pre-compensation (as in QwT) and post-compensation (in our QwT-v2), respectively. Note that in post-compensation, the compensation term $W X^\mathrm{quant} + b$ becomes $W Y^{\mathrm{quant}} + b$.

This simple change \emph{makes the compensation easier} since it bypasses the complex non-linearity from activation functions or attentions inside the Transformer block. More importantly, we \emph{no longer need to store} the block input during the inference stage for compensation, which is inevitable in QwT.

Then we move on and focus on another difficulty of QwT: the compensation matrix $W$ adds more parameters and latency, especially in low-bit quantization. More importantly, $WX^\mathrm{quant}+b$ is computed in floating-point format, rendering QwT incompatible with many AI chips. We need \emph{a lightweight and hardware-friendly} compensation method. 

By visualizing the distribution of (absolute) values in $W$ (with an example shown in Figure~\ref{fig:mot_b}), we find that the distribution is irregular in QwT's pre-compensation, but in our post-compensation the weights are \emph{concentrated on the diagonal}. Inspired by this observation, we hypothesize that \emph{using diagonal instead of full matrices can effectively compensate the error as well}, and call this idea channel-wise affine compensation (CWAC). Compared to the compensation module in QwT, CWAC is significantly more efficient in terms of parameter, memory-access, and compute. Furthermore, we will show that CWAC can be completely absorbed into existing modules by slightly modifying the inference engine (Section~\ref{ssec:integrate}). This means that our method \emph{inserts zero extra inference overhead and is compatible with existing AI chips}.

QwT asks $Y^{\mathrm{quant}}+W X^\mathrm{quant} + b$ to be close to $Y^{\mathrm{full}}$. In QwT-v2 we ask $W Y^{\mathrm{quant}} + b$, not $Y^{\mathrm{quant}} + W Y^{\mathrm{quant}} + b$, to be close to $Y^{\mathrm{full}}$. This key change is because now our compensation is not only post-compensation but also diagonal, hence we do not need the residual connection any more. That is, \emph{in Figure~\ref{fig:mot_a} the $\oplus$ should be removed and the two branches should become only one}.

\subsection{Preliminaries}

Before presenting details of our QwT-v2, we briefly outline basic concepts in quantization. Uniform quantization is probably the most popular method thanks to its simplicity and hardware-friendliness. Given a quantization bit-width $b$, uniform quantization turns floating-point tensor $x$ into:
\begin{equation}
    \label{eq:uniform_quant}
    x^\mathrm{quant} = \text{clip}\left( \nint{\frac{x}{s}} + z, 0, 2^b - 1 \right) \,,
\end{equation}
where $x^\mathrm{quant}$ is the quantized value of $x$, $\nint{\cdot}$ is the rounding function, and $\text{clip}(x,u,v)$ clips $x$ to the $[a,b]$ range. $s\in\mathbb{R}^{+}$ is the quantization scale and $z\in\mathbb{Z}$ is zero-point, which are determined by:
\begin{equation}
    s = \frac{\max(x) - \min(x)}{2^b - 1}\,, \quad z = \text{clip}\left( \nint{-\frac{\min(x)}{s}}, 0, 2^b - 1 \right) \,.
\end{equation}
Another commonly used quantization method is log2 quantization, which performs logarithmic compression on the input floating-point tensor before rounding and clipping. 

There are usually two ways of quantizing a tensor: per-tensor and per-channel. Per-tensor quantizes the whole tensor with only one $s$ and one $z$. Per-channel quantization quantizes the $i$-th channel with $s_i$ and $z_i$. Modern methods often apply per-channel quantization for weights and per-tensor quantization for activations, which achieves the balance between accuracy and efficiency.

\subsection{Channel-Wise Affine Compensation}
\label{ssec:cwac}
Given a floating-point model $B$ and its quantized counterpart $Q$, we propose Channel-Wise Affine Compensation (CWAC) to recover accuracy degradation caused by quantization. For \emph{each} quantized module in $Q$, we add a channel-wise affine transformation that aligns the quantized outputs with their floating-point counterparts.

Let $M_\mathrm{full}: \mathbb{R}^{C_\mathrm{in}} \rightarrow \mathbb{R}^{C_\mathrm{out}}$ denotes one module in the original network, and $M_\mathrm{quant}: \mathbb{R}^{C_\mathrm{in}} \rightarrow \mathbb{R}^{C_\mathrm{out}}$ its quantized version. Given calibration input $X_\mathrm{in}\in \mathbb{R}^{N\times C_\mathrm{in}}$ in which $N$ denotes the number of calibration samples, the output would be:
\begin{itemize}
    \item Floating-point output: $Y^{\mathrm{full}} = M_\mathrm{full}(X_\mathrm{in})$,
    \item Quantized output: $Y^{\mathrm{quant}} = M_\mathrm{quant}(X_\mathrm{in})$.
\end{itemize}
For each \emph{single} output channel $c \in \{1,...,C_\mathrm{out}\}$, we compute channel-wise compensation parameters by solving the \emph{one-dimensional} linear regression problem:
\begin{equation}
    \min_{\alpha_c, \beta_c} \|Y^{\mathrm{full}}_{:,c} - (\alpha_c \odot Y^{\mathrm{quant}}_{:,c} + \beta_c)\|^2 \,,
\end{equation}
where $\odot$ denotes element-wise multiplication. This leads to a very simple and easy-to-calculate closed-form solution:
\begin{align}
    \alpha_c &= \frac{\text{Cov}(Y^{\mathrm{full}}_{:,c}, Y^{\mathrm{quant}}_{:,c})}{\text{Var}(Y^{\mathrm{quant}}_{:,c})} \,, \\
    \beta_c &= \mathbb{E}[Y^{\mathrm{full}}_{:,c}] - \alpha_c \mathbb{E}[Y^{\mathrm{quant}}_{:,c}] \,,
\end{align}
in which $\text{Cov}$ and $\text{Var}$ are empirical covariance and variance, respectively. We get $\alpha\in\mathbb{R}^{C_\mathrm{out}}, \beta\in\mathbb{R}^{C_\mathrm{out}}$ by concatenating all $\alpha_c$ and $\beta_c$. The compensated output is then $\alpha \odot Y^{\mathrm{quant}} + \beta$.

\begin{figure}
    \centering
    \includegraphics[width=0.65\linewidth]{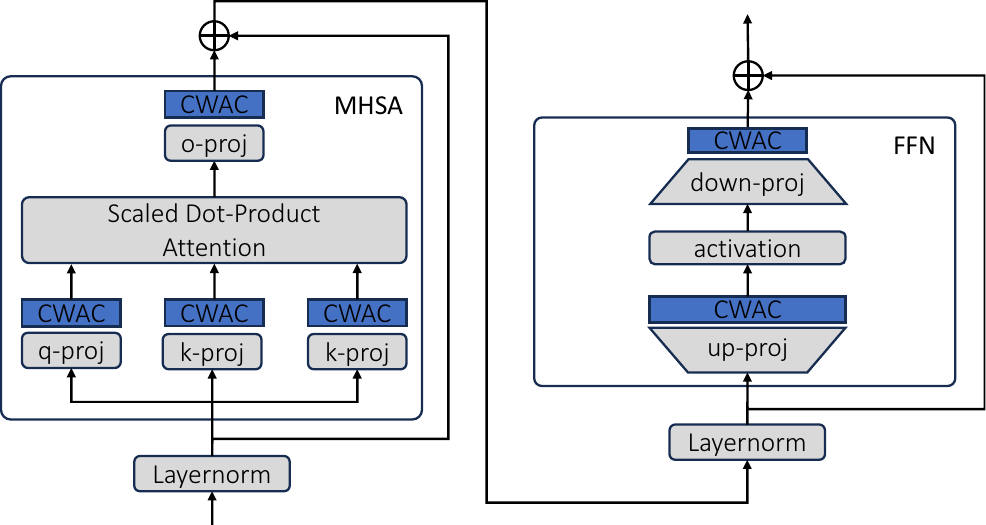}
    \caption{The architecture of QwT-v2's compensation in a vision Transformer. Compensations are applied after each fully connected layer.}
    \label{fig:cwac}
\end{figure}

In Figure~\ref{fig:cwac}, we illustrate our QwT-v2 compensation in a Transformer block. We add our CWAC (channel-wise affine compensation) modules after every quantized linear layer in Vision Transformer, and every quantized convolutional layer in CNN.

\subsection{Integrating into Inference Engine}
\label{ssec:integrate}

Although the compensation module in this form is already highly efficient, we will demonstrate that QwT-v2 can further eliminate the computational overhead by integrating parameters and computations into the quantize inference engine.

Here we discuss a potential integration framework following the representation in~\cite{Jacob05877}. Given a typical quantized linear layer $(W_q,b_q)$ and input $x_q$, where weights adopt per-channel quantization, and activations adopt per-tensor quantization, we have $(S_x, Z_x)$, $(S_W, Z_W)$, $(S_r, Z_r)$ for activations, weights and output, respectively ($S$ for quantization scale and $Z$ for zero-point). Note that $S_W,Z_W$ are $d$-dimensional vectors ($d$ denotes feature dimension), $S_x,S_r$ are floating-point numbers, $Z_x,Z_r$ are 32-bit integers. 
Now without considering bias $b$ (which is handled separately in an inference engine), we have $r=Wx$. In the quantized version, it is
\begin{equation}
\label{eq:quant-product-1}
S_r(r_q-Z_r) = S_W(W_q - Z_W) S_x(x_q - Z_x),
\end{equation}
which can be rewritten as
\begin{equation}
\label{eq:quant-product-2}
r_q = Z_r + M (W_qx_q-Z_W\odot x_q-Z_x W_q+Z_xZ_W),
\end{equation}
where the \emph{multiplier} $M$ is defined as
\begin{equation}
\label{eq:multiplier}
M = \frac{S_x S_W}{S_r}\in\mathbb{R}^d.
\end{equation}

Now, if we put our compensation module into consideration, we can get:
\begin{equation}
\label{eq:quant-product-1-comp}
S_r(r_q-Z_r) = \alpha S_W(W_q - Z_W) S_x(x_q - Z_x)+\beta,
\end{equation}
which can be rewritten as
\begin{equation}
\label{eq:quant-product-2_comp}
r_q = Z_r + M' (W_qx_q-Z_W\odot x_q-Z_x W_q+Z_xZ_W+\nint{\dfrac{\beta}{\alpha S_xS_W}}),
\end{equation}
\begin{equation}
\label{eq:multiplier_comp}
M' = \frac{\alpha S_x S_W}{S_r}\in\mathbb{R}^d.
\end{equation}
This integration procedure can be easily extended to other quantized layers (e.g., convolutional layer) and other hardware friendly quantization methods (e.g., log2 quantization). Notice that the calculation of Equation~\ref{eq:quant-product-2} is already efficiently implemented in inference engines (e.g., by \cite{tflite}), and Equation~\ref{eq:quant-product-2_comp} is also fully compatible with inference engine, with only numeric adjustments of $M$ and the offset. During this process, integration of $\alpha$ is lossless, and integration of $\beta$ is not lossless due to rounding errors.
But this information loss is negligible since the magnitude of $\frac{\beta}{\alpha S_xS_W}$ is usually large, which means the effect of rounding error is minimal (Table~\ref{tab:abl_rounding}).


\section{Experiments}

In this section, we evaluate our QwT-v2 on tasks in various domains, including visual classification, detection, generation, language and multimodal following the experimental setting of QwT (\cite{QwT}). We compare QwT-v2 with QwT and other quantization methods, plus ablation studies. 

\subsection{Experiments on Image classification}

\begin{table}
\centering
\small
\caption{Results (Top-1 accuracy) on ImageNet recognition. 
``\#-bit'' indicates bit-width for both weight and activation. 
``Size'' (MB) denotes model storage cost. ``Delta'' (MB) denotes extra storage cost from compensation modules.}
\label{tab:im_backbones}
\begin{tabular}{@{}ll@{\ }rrrrrr}
\toprule
\multirow{2}{*}{Network} & \multirow{2}{*}{Method} & \multicolumn{3}{c}{4-bit} & \multicolumn{3}{c}{6-bit} \\
\cmidrule(lr){3-5} \cmidrule(lr){6-8}
 & & \multicolumn{1}{c}{Size} & \multicolumn{1}{c}{Delta} & \multicolumn{1}{c}{Top-1} & \multicolumn{1}{c}{Size} & \multicolumn{1}{c}{Delta} & \multicolumn{1}{c}{Top-1} \\ 
\midrule

\multirow{4}{*}{Swin-T}  
 & Floating-point~\footnotesize{(32-bit)}          & 113.2 &  & 81.4 & \\ 
\cdashline{2-8}
 & RepQ-ViT (\cite{RepQ-ViT})               & 14.9      & -- & 73.0       & 21.7   & --    & 80.6       \\
 & RepQ-ViT + QwT (\cite{QwT})         & 19.2   & 4.3  & 75.5       & 26.0   & 4.3   & 80.7       \\
  & RepQ-ViT + QwT-v2         & 15.2   & 0.3   & \textbf{77.1}       & 22.0   & 0.3   & \textbf{80.9}       \\
\midrule

\multirow{4}{*}{ViT-B}   
 & Floating-point~\footnotesize{(32-bit)}          & 346.3 &  & 84.5 & \\ 
\cdashline{2-8}
 & RepQ-ViT (\cite{RepQ-ViT})               & 44.9   & --   & 68.5       & 66.2   & --   & 83.6       \\
 & RepQ-ViT + QwT         & 59.1      & 14.2 & 
 \textbf{76.3}       & 80.4  & 14.2    & \textbf{83.9}       \\
 & RepQ-ViT + QwT-v2         & 45.6   & 0.7   & 75.6       & 66.9   & 0.7   & 83.8       \\
\midrule

\multirow{4}{*}{ResNet-50} 
 & Floating-point~\footnotesize{(32-bit)}          & 102.2 & & 76.6 & \\ 
\cdashline{2-8}
 & Percentile (\cite{Percentile})             & 14.0 & --     & 68.4       & 19.9   & --   & 76.0       \\
 & Percentile  + QwT       & 16.0   & 2.0   & 74.5       & 21.9   & 2.0   & \textbf{76.8}       \\
 & Percentile  + QwT-v2       & 14.2   & 0.2   & \textbf{75.1}       & 20.1   & 0.2   & 76.7       \\
\bottomrule
\end{tabular}
\end{table}

We evaluated QwT-v2 on image classification on ImageNet (\cite{imagenet}), leveraging various backbone architectures including ViT (\cite{vit}), Swin Transformer (\cite{swin}) and ResNet (\cite{resnet}). Following the setting in QwT, we randomly sampled a calibration set with 512 samples from the ImageNet training set. We also performed $\beta$ rounding in the experiments. (if not specially noted, this setting was adopted in all following experiments).

We tested performance of QwT-v2 on various backbone, and the results are shown in Table~\ref{tab:im_backbones}. On Swin and ViT, we choose RepQ-ViT (\cite{RepQ-ViT}) as the baseline method, and Percentile (\cite{Percentile}) for ResNet. Compared to the backbone methods, QwT-v2 shows considerable improvements. We can find that the improvement on 4-bit quantized model (4-8\%) is greater than that on 6-bit. 

On Swin-T and ResNet, QwT-v2 outperforms QwT, but is slightly lower than QwT on ViT. In terms of extra parameters, QwT-v2 only accounts for about 1\%-2\% extra parameters compared to QwT's 30\%, which is a significant advantage for QwT-v2 in real-world deployment. Also note these extra parameters and computations can be merged (removed) if we integrate them into the inference engine.

\subsection{Experiments on Object Detection}

We evaluated our method on object detection and instance segmentation tasks using the COCO 2017 (\cite{COCO}) dataset. Following the setting in \cite{QwT}, we adopted Swin-S with Mask R-CNN (\cite{mask_rcnn}), and Swin-S/B with Cascade Mask R-CNN (\cite{cascade_r_cnn}) as detectors.

Results shown in Table~\ref{tab:detection} demonstrate QwT-v2's exceptional ability on detection model's compensation. QwT-v2 outperforms baseline methods and QwT on all settings. Also, the number of extra parameters and computations required by QwT-v2 is almost negligible compared to those in QwT.

\begin{table}
\centering
\small
\setlength{\tabcolsep}{2.4pt}
\caption{Detection results on COCO. \#Bits indicates the bit-width of weight/activation. We use box average precision ($\text{AP}^{\text{box}}$) and mask average precision ($\text{AP}^{\text{mask}}$) to assess object detection and instance segmentation accuracy, respectively.}
\label{tab:detection}
\setlength{\tabcolsep}{0.5mm}
\begin{tabular}{llcccc}
\toprule
Network                       & Method           & \#Bits       & Size       & $\text{AP}^{\text{box}}$    & $\text{AP}^{\text{mask}}$  \\ \hline
\multirow{7}{*}{\begin{tabular}[l]{@{}l@{}}\quad \quad Swin-S\\ + Mask R-CNN \end{tabular}}       & Floating-point                   & 32/32        & 276.5  & 48.5  & 43.3 \\ \cdashline{2-6}
                              & RepQ-ViT  &  4/4         & \phantom{0}36.1        & 42.6     & 40.0      \\ 
                              & RepQ-ViT + QwT             &  4/4         & \phantom{0}44.0        & 43.1     & 40.4      \\ 
                              & RepQ-ViT + QwT-v2             &  4/4         & \phantom{0}36.7        & \textbf{43.4}     & \textbf{40.5}      \\ 
                              \cdashline{2-6}
                              & RepQ-ViT   &  6/6         & \phantom{0}53.3        & 47.6     & 42.9      \\ 
                              & RepQ-ViT + QwT             &  6/6         & \phantom{0}61.2        & \textbf{48.0}     & \textbf{43.1}      \\
                              & RepQ-ViT + QwT-v2             &  6/6         & \phantom{0}53.9        & \textbf{48.0}     & \textbf{43.1}      \\
                              \hline
\multirow{7}{*}{\begin{tabular}[l]{@{}l@{}}\quad \ \ Swin-S\\ \ \ \ + Cascade \\ \ \ Mask R-CNN \end{tabular}}       & Floating-point                   & 32/32        & 427.8  & 51.9  & 45.0 \\ \cdashline{2-6}
                              & RepQ-ViT   &  4/4         & \phantom{0}56.9        & 49.3     & 43.1      \\ 
                              & RepQ-ViT + QwT             &  4/4         & \phantom{0}64.8        & 49.9     & 43.4      \\ 
                              & RepQ-ViT + QwT-v2             &  4/4         & \phantom{0}57.5        & \textbf{50.0}     & \textbf{43.6}      \\ 
                              \cdashline{2-6}
                              & RepQ-ViT   &  6/6         & \phantom{0}83.4        & 51.4     & 44.6      \\ 
                              & RepQ-ViT + QwT             &  6/6         & \phantom{0}91.3        & 51.7     &  44.8      \\
                              & RepQ-ViT + QwT-v2             &  6/6         & \phantom{0}84.0        & \textbf{52.0}     &  \textbf{45.0}      \\
                              \hline
\multirow{7}{*}{\begin{tabular}[l]{@{}l@{}}\quad \ \ Swin-B\\ \ \ \ + Cascade \\ \ \ Mask R-CNN \end{tabular}}       & Floating-point                   & 32/32        & 579.9  & 51.9  & 45.0 \\ \cdashline{2-6}
                              & RepQ-ViT   &  4/4         & \phantom{0}76.1         & 49.3     & 43.1      \\ 
                              & RepQ-ViT + QwT             &  4/4         & \phantom{0}90.1         & 50.0     & 43.7      \\ 
                              & RepQ-ViT + QwT-v2             &  4/4         & \phantom{0}77.0         & \textbf{50.3}     & \textbf{43.8}      \\ 
                              \cdashline{2-6}
                              & RepQ-ViT   &  6/6         & 112.1        & 51.5     & 44.8      \\ 
                              & RepQ-ViT + QwT             &  6/6         & 126.1        & 51.8     & 45.0      \\
                              & RepQ-ViT + QwT-v2             &  6/6         & 113.0        & \textbf{51.9}     & \textbf{45.1}      \\
\bottomrule
\end{tabular}
\end{table}

\subsection{Experiments on Multimodal Recognition}

We conducted experiments on CLIP (\cite{CLIP}). We test our QwT-v2 on CLIP to verify its ability on multimodal tasks. Following QwT (\cite{QwT}), we chose ViT-B/32 as the visual encoder and a 12-block Transformer as the text encoder. We randomly selected 512 image-text pairs from the training data for baseline calibration and QwT-v2 initialization.

Results in Table~\ref{tab:clip} show that QwT-v2 is also very effective on multimodal model's compensation. QwT-v2 improves top-1 accuracy over all baselines on all settings. And we can find out that the compensation improvement of compensating both vision and text encoder is greater than compensating vision encoder only. On (vision only) quant setup, result of QwT-v2 is slightly behind QwT on 6/6-bit quantization but slightly better than QwT on 8/8-bit quantization. On (vision \& text) setting, QwT-v2 outperforms QwT by a large margin on both bit-widths (1.8\% on 6/6-bit and 1.9\% on 8/8-bit) with significantly less extra parameters.

\subsection{Experiments on Image Generation}

We also conducted experiments on quantized diffusion model, to evaluate the generated image quality from full precision DiT-XL/2 model, quantized model, and compensated model, under 256x256 resolution. We utilized the DDIM sampler with 50 sampling steps and applied classifier-free guidance (cfg) of 1.5 to realize between efficiency and precision. We also followed the key assumption from~\cite{QwT}: quantization error mainly depend on input token $x$, and is minimally affected by timestamp $t$. We find this assumption practical and chose $t=0$ for compensation initialization.

We built our QwT-v2 on Q-DiT quantized model. As shown in Table~\ref{tab:dit}, QwT-v2 shows competitive results compared to other PTQ methods and QwT. On W8A8 quantization, FID of QwT is 0.03 higher than QwT, but shows better inception score compared with QwT. On W4A8 quantization, QwT-v2 demonstrates both the best FID score and and inception score, with 0.34 advantage on FID and 18.4 advantage on IS compared to second best results from QwT.

\begin{table}[t]
\centering
\small
\begin{minipage}[t]{0.5\textwidth}
\centering
\setlength{\tabcolsep}{1.5pt}
\caption{Quantization results of CLIP for zero-shot classification tasks on ImageNet. Note that the `Setup' column differentiates between two strategies: quantizing only the Visual Encoder, and quantizing both the Visual and Text Encoders concurrently.}
\begin{tabular}{llccc}
\toprule
Setup                       & Method                           & \#Bits       & Size (MB)       & Top-1  \\ \hline
\multirow{7}{*}{Vision}       & Floating-point                  & 32/32             & 607.2       & 63.4  \\ \cdashline{2-5}
                              & RepQ-ViT        &  6/6              & 323.5       & 59.2  \\ 
                              & RepQ-ViT + QwT                  &  6/6              & 336.8       & \textbf{60.3}  \\ 
                              & RepQ-ViT + QwT-v2                  &  6/6              & 324.2       & 59.8  \\ 
                              \cdashline{2-5}
                              & RepQ-ViT        &  8/8              & 345.3       & 62.9  \\ 
                              & RepQ-ViT + QwT                  &  8/8              & 359.5       & 63.0  \\
                              & RepQ-ViT + QwT-v2                  &  8/8              & 346.0       & \textbf{63.1}  \\
                              \hline
\multirow{7}{*}{\begin{tabular}[l]{@{}c@{}}Vision  \\ \& Text \end{tabular}}       & Floating-point                   & 32/32        & 607.2  & 63.4  \\ \cdashline{2-5}
                              & RepQ-ViT        &  6/6             & 200.8        & 29.8      \\ 
                              & RepQ-ViT + QwT                  &  6/6             & 221.3        & 43.5      \\ 
                              & RepQ-ViT + QwT-v2                 &  6/6             & 202.0        & \textbf{45.3}      \\ 
                              \cdashline{2-5}
                              & RepQ-ViT        &  8/8             & 232.1        & 38.7      \\ 
                              & RepQ-ViT + QwT                  &  8/8             & 252.6        & 54.6      \\
                              & RepQ-ViT + QwT-v2                  &  8/8             & 233.3        & \textbf{56.5}      \\

\bottomrule
\end{tabular}
\label{tab:clip}
\end{minipage}
\hfill
\begin{minipage}[t]{0.48\textwidth}
    \centering
    \caption{Results of quantized DiT-XL/2 model. \#Bits indicates the bit-width of weight/activation. We use FID and Inception Score (IS) to measure generated image quality. $\downarrow$ ($\uparrow$) means smaller (larger) is better.}
	\setlength{\tabcolsep}{1pt}
    \renewcommand{\arraystretch}{1.2}
    \begin{tabular}{lcccc}
        \toprule
        Method & \#Bits & Size (MB) & \phantom{0}FID ($\downarrow$) & IS ($\uparrow$) \\
        \midrule
        Floating-point & 16/16  & 1349 & \phantom{00}5.32 & 236.17 \\
        \midrule
        RepQ-ViT & 8/8 & \phantom{0}677 & \phantom{00}5.46 & 234.74 \\
        GPTQ & 8/8 & \phantom{0}690 & \phantom{00}5.90 & 218.90 \\
        Q-DiT& 8/8 & \phantom{0}683 & \phantom{00}5.45 & 236.52 \\
        Q-DiT + QwT &8/8 & \phantom{0}707 & \phantom{00}\textbf{5.35} & 236.91 \\
        Q-DiT + QwT-v2 &8/8 & \phantom{0}685 & \phantom{00}5.38 & \textbf{244.34} \\
        \midrule
        RepQ-ViT & 4/8 & \phantom{0}339 & 319.68 & \phantom{00}2.20 \\
        GPTQ & 4/8 & \phantom{0}351 & \phantom{00}9.94 & 166.35 \\
        Q-DiT & 4/8 & \phantom{0}347 & \phantom{00}6.75 & 208.38 \\
        Q-DiT + QwT & 4/8 & \phantom{0}361 & \phantom{00}6.06 & 215.70 \\
        Q-DiT + QwT-v2 & 4/8 & \phantom{0}349 & \phantom{00}\textbf{5.72} & \textbf{234.06} \\
        \bottomrule
    \end{tabular}
    \label{tab:dit}
\end{minipage}
\end{table}

\subsection{Experiments on Large Language Models}

We evaluated our framework on LLaMA3-8B (\cite{dubey2024llama3herdmodels}). For PTQ methods, we adopted GPTQ (\cite{gptq}) with INT4 weight quantization only. We conducted a group-wise asymmetric quantization with a group size of 128 and applied activation reordering. In particular, GPTQ take 128 samples from the C4 dataset as calibration sets, and each sample is 2048 tokens long. Like QwT, we use exactly the same calibration set for QwT-v2.

Following the settings of GPTQ, we evaluated the perplexity on the WikiText2 (\cite{stephen2017pointer}) and C4 (\cite{raffel2020exploring}) datasets. We further assessed the zero-shot commonsense question answering (QA) ability on eight tasks which are evaluated by QwT (\cite{QwT}), including SIQA (\cite{sap2019social}), HellaSwag (\cite{zellers2019hellaswag}), PIQA (\cite{Bisk2020piqa}), WinoGrande (\cite{sakaguchi2021winogrande}), ARC (\cite{clark2018think}), BoolQ (\cite{clark2019boolq}), and OpenBookQA (\cite{mihaylov2018can}). 


Table~\ref{tab:llm_ptq_method} summarizes the perplexity in WikiText2 (W2), C4 and the average accuracy in eight common sense reasoning datasets. Our QwT-v2 achieves consistent gains over baseline the method GPTQ, and demonstrates close performance compared to QwT.

\begin{table}[t]
    \centering
    \small
    \setlength{\tabcolsep}{3pt}
    \caption{Quantization results among WikiText2 (W2), C4 and eight zero-shot commonsense QA datasets using LLaMA3-8B as the backbone. $\downarrow$ ($\uparrow$) means smaller (larger) is better.}
    \begin{tabular}{lccccc}
    \toprule
    Method & \#Bits & Size (GB) &  W2 ($\downarrow$) & C4 ($\downarrow$) & QA. Avg ($\uparrow$)     \\ \hline
    Floating-point     & 16  & 16.06 & 6.24 & 8.96 & 66.10 \\ \hline
    GPTQ    & 4   & \phantom{0}5.73 & 6.65 & 9.44 &  64.90  \\
    GPTQ + QwT     & 4   & \phantom{0}6.80 & 6.63 & \textbf{9.38}  & 65.18\\ 
    GPTQ + QwT-v2     & 4   & \phantom{0}5.74 & \textbf{6.62} & 9.40  & \textbf{65.21}\\ 
    \bottomrule
    \end{tabular}
    \label{tab:llm_ptq_method}
    \end{table}

\subsection{Ablation studies}

we conducted ablation studies on QwT-v2 to verify the effectiveness of its design choices.

\textbf{Ablation on varying calibration set size.} We changed the size of calibration set in QwT-v2, and report the result on ViT-B and Swin-T quantized by RepQ-ViT. As shown in Table~\ref{tab:abl_size}, by increasing calibration set size, the result can get slightly better. On ViT-B, the results are more significantly affected (0.5\% gap between 32 images and 1024 images), while on Swin-T, the results are less affected (0.2\% gap between 32 images and 1024 images). Choosing a calibration set with 512 images can achieve the balance between efficiency and effectiveness.

\textbf{Ablation on compensation position.} To assess the effectiveness of fine-grained compensation (compensating behind each quantized linear/conv layer). We conduct ablation experiment on only compensating the final fc/conv in each block, results are show in Table~\ref{tab:abl_position} (ViT-B and Swin-T are quanted by RepQ-ViT, Resnet-50 by Percentile). From the table, we can summarize that our fine-grained compensation (`all') consistently provide a gain compared to only compensating final fc/conv (`post'). And the gain in 4-bit quantization is always greater than 6-bit quantization. The results clearly demonstrated the effectiveness of fine-grained compensation in QwT-v2.

\textbf{Ablation on rounding $\beta$.} We further investigated whether rounding $\beta$ to integers will cause performance drop, and results are shown in Table~\ref{tab:abl_rounding}. We can see that there is nearly no gap between rounded $\beta$'s results and its floating-point counterparts.

\begin{table}[t]
\centering
\small
\setlength{\tabcolsep}{4pt}
\caption{Ablation on varying the size of calibration set in QwT-v2 on 4/4-bit quantized ViT-B and Swin-T model. `Top-1' means Top-1 accuracy on ImageNet validation set.}
\begin{tabular}{llccc}
\toprule
Network  & Calibration Size   & Top-1\\ \hline
\multirow{4}{*}{\begin{tabular}[l]{@{}c@{}}ViT-B  \\  (4/4-bit) \end{tabular}}
& 32 images & 75.2\\ 
& 128 images & 75.3 \\ 
& 512 images & 75.6 \\ 
& 1024 images & 75.7 \\ 
\hline
\multirow{4}{*}{\begin{tabular}[l]{@{}c@{}}Swin-T  \\  (4/4-bit) \end{tabular}}       
& 32 images & 69.9\\ 
& 128 images & 77.0 \\ 
& 512 images & 77.1 \\ 
& 1024 images & 77.1 \\ 
\bottomrule
\end{tabular}
\label{tab:abl_size}
\end{table}

\begin{table}[t]
\centering
\small
\begin{minipage}[t]{0.48\textwidth}
\centering
\setlength{\tabcolsep}{4pt}
\caption{Ablation on changing the position of compensation modules. `all' denotes compensating behind all quant linear/conv layers. `post' denotes only compensate behind final fc/conv.}
\begin{tabular}{llccc}
\toprule
Network  & \#Bits & Position  & Top-1\\ \hline
\multirow{4}{*}{ViT-B}
& 4/4 & post & 73.5\\ 
& 4/4 & all & 75.6 \\ 
\cdashline{2-4}[1pt/1pt]
& 6/6 & post & 83.5\\ 
& 6/6 & all & 83.8 \\ 
\hline
\multirow{4}{*}{Swin-T}       
& 4/4 & post & 73.9 \\ 
& 4/4 & all & 77.1\\ 
\cdashline{2-4}[1pt/1pt]
& 6/6 & post & 80.6 \\ 
& 6/6 & all & 80.9 \\ 
\hline
\multirow{4}{*}{Resnet-50}       
& 4/4 & post & 73.1 \\ 
& 4/4 & all & 75.1 \\ 
\cdashline{2-4}[1pt/1pt]
& 6/6 & post & 76.5 \\ 
& 6/6 & all & 76.7 \\ 
\bottomrule
\end{tabular}
\label{tab:abl_position}
\end{minipage}
\hfill
\begin{minipage}[t]{0.48\textwidth}
\centering
\setlength{\tabcolsep}{4pt}
\caption{Ablation on the effect of rounding $\beta$ as described in Sec.~\ref{ssec:integrate}. The $\checkmark$ on the `Round' column denotes the result of rounding $\beta$, which is by default adopted in all the experiments.}
\begin{tabular}{llccc}
\toprule
Network  & \#Bits & Round  & Top-1\\ \hline
\multirow{4}{*}{ViT-B}
& 4/4 & & 75.6 \\ 
& 4/4 & $\checkmark$ & 75.6 \\ 
\cdashline{2-4}[1pt/1pt]
& 6/6 & & 83.8 \\ 
& 6/6 & $\checkmark$ & 83.8 \\ 
\hline
\multirow{4}{*}{Swin-T}       
& 4/4 & & 77.1 \\ 
& 4/4 & $\checkmark$ & 77.1 \\ 
\cdashline{2-4}[1pt/1pt]
& 6/6 & & 90.0 \\ 
& 6/6 & $\checkmark$ & 80.9 \\ 
\hline
\multirow{4}{*}{ViT-B}       
& 4/4 & & 75.3 \\ 
& 4/4 & $\checkmark$ & 75.1 \\ 
\cdashline{2-4}[1pt/1pt]
& 6/6 & & 76.7 \\ 
& 6/6 & $\checkmark$ & 76.7 \\ 
\bottomrule
\end{tabular}
\label{tab:abl_rounding}
\end{minipage}
\end{table}

\section{Conclusions}

In this paper, we proposed QwT-v2, which addresses major limitations issues in QwT, including: low hardware compatibility, massive additional parameters, and increased inference latency. These are realized by two key insights based on QwT. First, we change the pre-compensation in QwT to post-compensation, which lowers inference overhead and frees us from learning the complex non-linearity within a Transformer block. Consequently, we can move on to further replace the linear compensation matrices with diagonal ones, which saves parameters and computations to a large extent. More importantly, this compensation module can be integrated into inference engine with minimal effort, and completely eliminate extra parameters and computations. The proposed QwT-v2 has exhibited both low parameter and computation costs and high accuracy compared to QwT in various application: recognition, detection, multimodal, LLM, and image generation.

\section{Limitations and Future Work}

There are some intriguing drawbacks which are not fully discussed and utilized in QwT-v2. We find that compensation from earlier layers are more prominent and more effective than compensation from deeper layers, and compensation for different modules have different effectiveness. It is worth studying how to make those less effective compensation positions work better.

Another limitation of QwT-v2 is that further finetuning the compensation parameters by 1 epoch is not as effective as that operation in QwT. This is not surprising, as the extra parameter size in QwT-v2 is only around 5-10\% of that in QwT. In future work, we are interesting in finding more innovative ways to improve accuracy after finetuning while still maintaining parameter and computation efficiency.

Finally, QwT-v2 and QwT are both built upon an existing PTQ method, which means we need to develop better PTQ methods in areas where they work not so well, e.g., when quantizing both activations and weights in an LLM.



\newpage
\appendix

\section{Implementation Details}

In this section, we present full implementation details of the different types of tasks in our experiments.

\textbf{Image Classification.} Following QwT, We selected RepQ-ViT~\cite{RepQ-ViT} for ViT quantization, and Percentile~\cite{Percentile} for ResNet, as the primary baseline PTQ methods to integrate with QwT-v2 method. Following~\cite{RepQ-ViT}, we randomly sampled 32 images from the ImageNet~\cite{imagenet} dataset as the calibration set to initialize the quantized weights for these baseline methods. Additionally, a separate set of 512 randomly selected images from the ImageNet training set was used to initialize the parameters of the QwT modules (excluding PTQ weights). For all networks, the $W$ matrices are in FP32 format (unlike QwT, which set compesantion matrices to FP16 to reduce model size, the size of QwT-v2 compensation matrices are extremely small, so FP32 works well here), and $b$ is rounded to INT32 format by default.

All the experiments above on image classification are implemented on a single RTX 3090 GPU with 24GB memory.

Additionally, When fine-tuning the QwT modules along with the classification head for an additional epoch, following \cite{QwT}, we utilized AdamW~\cite{adamw} as the optimizer. The batch size was set to 32 per GPU (using a total of 4 GPUs), and weight decay was set to 0. The learning rate was configured to 1e-7 for ViT~\cite{vit}, 5e-6 for Swin~\cite{swin}, 1e-5 for ResNet~\cite{resnet}. 

\textbf{Object Detection \& Instance Segmentation.} Following~\cite{RepQ-ViT}, we randomly sampled a single image from the COCO dataset~\cite{COCO} to initialize the quantized weights for baseline PTQ methods. All other details are consistent with the image classification case. Experiments shown in \ref{tab:detection} are also implemented on a single RTX 3090 GPU with 24GB memory.

\textbf{Image Generation.} Consistent with the experimental setup of Q-DiT~\cite{Q-DiT}, we selected the DiT architecture and employed pretrained DiT-XL/2 models at a resolution of $256 \times 256$. Experiments are implemented on 4 RTX 3090 GPUs.

\textbf{Large Language Models.} Experiments on large language models follow the baseline of GPTQ~\cite{gptq}, and we choose Llama3-8B as our backbone.the computation. the initialization of QwT-v2 is done on 4 RTX 3090 GPUs.

All the experiments are implemented with pyTorch.

\newpage
\section{More Results on Image Recognition}

We further show results on more backbone in Table~\ref{tab:more_im_backbones}. We extend the results to DeiT-Tiny, DeiT-Small, Swin-Tiny, Swin-Small, ViT-B, ResNet-50 and ResNet-101. Results shows that QwT-v2 works well on all kinds of backbone, especially on 4-bit quantized backbones.

\begin{table}[ht]
\centering
\small
\caption{Results (Top-1 accuracy) on ImageNet recognition with more backbones. 
``\#-bit'' indicates bit-width for both weight and activation. 
``Size'' (MB) denotes model storage cost. ``Delta'' (MB) denotes extra storage cost from compensation modules.}
\label{tab:more_im_backbones}
\begin{tabular}{@{}ll@{\ }rrrrrr}
\toprule
\multirow{2}{*}{Network} & \multirow{2}{*}{Method} & \multicolumn{3}{c}{4-bit} & \multicolumn{3}{c}{6-bit} \\
\cmidrule(lr){3-5} \cmidrule(lr){6-8}
 & & \multicolumn{1}{c}{Size} & \multicolumn{1}{c}{Delta} & \multicolumn{1}{c}{Top-1} & \multicolumn{1}{c}{Size} & \multicolumn{1}{c}{Delta} & \multicolumn{1}{c}{Top-1} \\ 
\midrule

\multirow{4}{*}{DeiT-T}  
 & Floating-point~\footnotesize{(32-bit)}          & 22.9 &  & 72.2 & \\ 
\cdashline{2-8}
 & RepQ-ViT (\cite{RepQ-ViT})               & 3.3      & -- & 58.2       & 4.6   & --    & 71.0       \\
 & RepQ-ViT + QwT (\cite{QwT})         & 4.2   & 0.9  & \textbf{61.4}       & 5.5   & 0.9   & \textbf{71.2}       \\
  & RepQ-ViT + QwT-v2         & 3.4   & 0.1   & 59.9       & 4.7   & 0.1   & \textbf{71.2}       \\
\midrule

\multirow{4}{*}{DeiT-S}  
 & Floating-point~\footnotesize{(32-bit)}          & 88.2 &  & 79.9 & \\ 
\cdashline{2-8}
 & RepQ-ViT          & 11.9      & -- & 69.0       & 17.2   & --    & 78.9       \\
 & RepQ-ViT + QwT     & 15.4   & 3.5  & \textbf{71.5}       & 20.7   & 3.5   & \textbf{79.1}       \\
  & RepQ-ViT + QwT-v2    & 12.2   & 0.3   & 71.1       & 17.5   & 0.3   & 78.9       \\
\midrule

\multirow{6}{*}{Swin-T}  
 & Floating-point~\footnotesize{(32-bit)}          & 113.2 &  & 81.4 & \\ 
\cdashline{2-8}
 & RepQ-ViT           & 14.9      & -- & 73.0       & 21.7   & --    & 80.6       \\
 & RepQ-ViT + QwT      & 19.2   & 4.3  & 75.5       & 26.0   & 4.3   & 80.7       \\
  & RepQ-ViT + QwT-v2         & 15.2   & 0.3   & \textbf{77.1}       & 22.0   & 0.3   & \textbf{80.9}       \\
\midrule

\multirow{4}{*}{Swin-S}  
 & Floating-point~\footnotesize{(32-bit)}          & 198.4 &  & 83.2 & \\ 
\cdashline{2-8}
 & RepQ-ViT            & 25.8      & -- & 80.2       & 38.0   & --    & 82.8       \\
 & RepQ-ViT + QwT      & 33.7   & 7.9  & \textbf{80.4}       & 45.9   & 7.9   & 82.9       \\
  & RepQ-ViT + QwT-v2         & 26.4   & 0.6   & 80.3       & 38.6   & 0.6   & \textbf{83.0}       \\
\midrule

\multirow{4}{*}{ViT-B}   
 & Floating-point~\footnotesize{(32-bit)}          & 346.3 &  & 84.5 & \\ 
\cdashline{2-8}
 & RepQ-ViT           & 44.9   & --   & 68.5       & 66.2   & --   & 83.6       \\
 & RepQ-ViT + QwT         & 59.1      & 14.2 & 
 \textbf{76.3}       & 80.4  & 14.2    & \textbf{83.9}       \\
 & RepQ-ViT + QwT-v2         & 45.6   & 0.7   & 75.6       & 66.9   & 0.7   & 83.8       \\
\midrule

\multirow{4}{*}{ResNet-50} 
 & Floating-point~\footnotesize{(32-bit)}          & 102.2 & & 76.6 & \\ 
\cdashline{2-8}
 & Percentile (\cite{Percentile})             & 14.0 & --     & 68.4       & 19.9   & --   & 76.0       \\
 & Percentile  + QwT       & 16.0   & 2.0   & 74.5       & 21.9   & 2.0   & \textbf{76.8}       \\
 & Percentile  + QwT-v2       & 14.2   & 0.2   & \textbf{75.1}       & 20.1   & 0.2   & 76.7       \\
\midrule

\multirow{4}{*}{ResNet-101} 
 & Floating-point~\footnotesize{(32-bit)}          & 178.2 & & 77.3 & \\ 
\cdashline{2-8}
 & Percentile             & 23.7 & --     & 74.7       & 34.3   & --   & 77.1       \\
 & Percentile  + QwT       & 28.0   & 4.3   & \textbf{76.4}       & 38.6   & 4.3   & \textbf{77.2}       \\
 & Percentile  + QwT-v2       & 24.1   & 0.4   & 76.2 & 34.7   & 0.4  &  \textbf{77.2}     \\
\bottomrule
\end{tabular}
\end{table}

\newpage

We also show results of QwT-v2 on ImageNet classification when the compensation modules and classfication head are fine-tuned on ImageNet training set for one epoch. As shown in Table~\ref{tab:im_backbones_one_epoch}, the compensation accuracy of QwT-v2 were further improved when fine-tuned. But the margin of improvement is not as large as the improvement in QwT. This is not surprising, because the extra parameter account in QwT-v2 is only about 5-10\% to that in QwT. By the way, people may prefer not to fint-tune the compensation in most situations, because initial compensation of QwT-v2 only takes about 2-3 minutes, but fine-tuning on ImageNet requires several hours and significant more computation resources.

\begin{table}[ht]
\centering
\small
\caption{Results with one-epoch fine-tuning on ImageNet recognition. 
``\#-bit'' indicates bit-width for both weight and activation. 
``Size'' (MB) denotes model storage cost. ``Delta'' (MB) denotes extra storage cost from compensation modules.  `*' denotes
compensation modules and classification head are fine-tuned for one epoch on ImageNet training set.}
\label{tab:im_backbones_one_epoch}
\begin{tabular}{@{}ll@{\ }rrrrrr}
\toprule
\multirow{2}{*}{Network} & \multirow{2}{*}{Method} & \multicolumn{3}{c}{4-bit} & \multicolumn{3}{c}{6-bit} \\
\cmidrule(lr){3-5} \cmidrule(lr){6-8}
 & & \multicolumn{1}{c}{Size} & \multicolumn{1}{c}{Delta} & \multicolumn{1}{c}{Top-1} & \multicolumn{1}{c}{Size} & \multicolumn{1}{c}{Delta} & \multicolumn{1}{c}{Top-1} \\ 
\midrule

\multirow{6}{*}{Swin-T}  
 & Floating-point~\footnotesize{(32-bit)}          & 113.2 &  & 81.4 & \\ 
\cdashline{2-8}
 & RepQ-ViT               & 14.9      & -- & 73.0       & 21.7   & --    & 80.6       \\
 & RepQ-ViT + QwT         & 19.2   & 4.3  & 75.5       & 26.0   & 4.3   & 80.7       \\
  & RepQ-ViT + QwT*         & 19.2   & 4.3  & 79.3       & 26.0   & 4.3   & 80.9       \\
  & RepQ-ViT + QwT-v2         & 15.2   & 0.3   & 77.1       & 22.0   & 0.3   & 80.9       \\
  & RepQ-ViT + QwT-v2*         & 15.2   & 0.3   & 78.1       & 22.0   & 0.3   & 80.9       \\
\midrule

\multirow{6}{*}{ViT-B}   
 & Floating-point~\footnotesize{(32-bit)}          & 346.3 &  & 84.5 & \\ 
\cdashline{2-8}
 & RepQ-ViT               & 44.9   & --   & 68.5       & 66.2   & --   & 83.6       \\
 & RepQ-ViT + QwT         & 59.1      & 14.2 & 
 76.3       & 80.4  & 14.2    & 83.9       \\
  & RepQ-ViT + QwT*         & 59.1      & 14.2 & 
 78.5       & 80.4  & 14.2    & 84.0       \\
 & RepQ-ViT + QwT-v2         & 45.6   & 0.7   & 75.6       & 66.9   & 0.7   & 83.8       \\
  & RepQ-ViT + QwT-v2*         & 45.6   & 0.7   & 77.5       & 66.9   & 0.7   & 83.8       \\

\bottomrule
\end{tabular}
\end{table}

We also show results for QwT-v2 with PTQ4ViT~\cite{PTQ4ViT} as baseline method. As shown in Table~\ref{tab:im_backbones_ptq4vit}, QwT-v2 improves the accuracy compared to baseline by 25.6\% on 4-bit quantization and 0.6\% on 6-bit quantization.

\begin{table}[ht]
\centering
\small
\caption{Results with PTQ4ViT~\cite{PTQ4ViT} as baseline method. 
``\#-bit'' indicates bit-width for both weight and activation. 
``Size'' (MB) denotes model storage cost. ``Delta'' (MB) denotes extra storage cost from compensation modules.}
\label{tab:im_backbones_ptq4vit}
\begin{tabular}{@{}ll@{\ }rrrrrr}
\toprule
\multirow{2}{*}{Network} & \multirow{2}{*}{Method} & \multicolumn{3}{c}{4-bit} & \multicolumn{3}{c}{6-bit} \\
\cmidrule(lr){3-5} \cmidrule(lr){6-8}
 & & \multicolumn{1}{c}{Size} & \multicolumn{1}{c}{Delta} & \multicolumn{1}{c}{Top-1} & \multicolumn{1}{c}{Size} & \multicolumn{1}{c}{Delta} & \multicolumn{1}{c}{Top-1} \\ 
\midrule

\multirow{3}{*}{ViT-B}   
 & Floating-point~\footnotesize{(32-bit)}          & 346.3 &  & 84.5 & \\ 
\cdashline{2-8}
 & PTQ4ViT-ViT~(\cite{PTQ4ViT})               & 44.9   & --   & 30.7       & 66.2   & --   & 81.7       \\
 & PTQ4ViT-ViT+QwT-v2 & 45.6   & 0.7   & 56.3       & 66.9   & 0.7   & 82.3       \\
\bottomrule
\end{tabular}
\end{table}

\end{document}